\documentclass[preprint,12pt]{elsarticle}

\usepackage{amssymb}
\usepackage{amsmath}
\usepackage[utf8]{inputenc} 
\usepackage[T1]{fontenc}    
\usepackage{hyperref}       
\usepackage{url}            
\usepackage{booktabs}       
\usepackage{amsfonts}       
\usepackage{nicefrac}       
\usepackage{microtype}      
\usepackage{xcolor}         
\usepackage{mathtools}
\usepackage{amsthm}
\usepackage{ bbold }
\usepackage{float} 
\usepackage{wrapfig}
\usepackage{enumitem}
\usepackage{subfig}
\usepackage[capitalize,noabbrev]{cleveref}
\usepackage{bibunits}
\usepackage{xcolor}

\journal{Information Sciences}

\def\R{\mathbb{R}}
\def\exp{\mathrm{exp}}
\def\gnn{\mathrm{gnn}}
\def\MLP{\mathrm{MLP}}
\def\score{\mathrm{score}}

\def\RFM{\mathrm{RFM}}
\def\E{\mathrm{E}}

\def\ftc{FTC}

\def\ebeta{\beta_E}
\def\ealpha{\alpha_E}
\def\changed{}

\begin{document}

\begin{frontmatter}


\title{Diverse and Feasible Retrosynthesis using GFlowNets}


\author[label1,label2]{Piotr Gaiński\corref{cor1}}
\author[label3,label4]{Michał Koziarski}
\author[label5]{Krzysztof Maziarz}
\author[label5]{Marwin Segler}
\author[label1]{Jacek Tabor}
\author[label1]{Marek Śmieja}

\cortext[cor1]{piotr.gainski@doctoral.uj.edu.pl}

\affiliation[label1]{organization={Jagiellonian University},
            city={Kraków},
            country={Poland}}
\affiliation[label2]{organization={Jagiellonian University, Doctoral School of Exact and Natural Sciences},
            city={Kraków},
            country={Poland}}
\affiliation[label3]{organization={Mila – Québec AI Institute},
            city={Montréal},
            country={Canada}}
\affiliation[label4]{organization={Université de Montréal},
city={Montréal},
country={Canada}}
\affiliation[label5]{organization={Microsoft Research AI for Science},
city={Cambridge},
country={United Kingdom}}



\begin{abstract}
Single-step retrosynthesis aims to predict a set of reactions that lead to the creation of a target molecule, which is a crucial task in molecular discovery. Although a target molecule can often be synthesized with multiple different reactions, it is not clear how to verify the feasibility of a reaction, because the available datasets cover only a tiny fraction of the possible solutions. Consequently, the existing models are not encouraged to explore the space of possible reactions sufficiently. In this paper, we propose a novel single-step retrosynthesis model, RetroGFN, that can explore outside the limited dataset and return a diverse set of feasible reactions by leveraging a feasibility proxy model during the training. We show that RetroGFN achieves competitive results on standard top-k accuracy while outperforming existing methods on round-trip accuracy. Moreover, we provide empirical arguments in favor of using round-trip accuracy\changed{,} which expands the notion of feasibility with respect to the standard top-k accuracy metric.
\end{abstract}




\end{frontmatter}



\section{Introduction}
\label{sec:introduction}
The rising interest in machine learning \changed{has} led to the development of many deep generative models for de novo drug design 
\cite{gomez2018automatic,maziarz2021learning}. 
Such approaches can propose novel molecules with promising properties (e.g.\changed{,} high binding affinity score) predicted by other machine learning models 
\cite{walters2020applications,gainski2023chienn,chithrananda2020chemberta}
, however, these virtual compounds eventually need to be synthesized and evaluated in the wet lab. This motivates the development of reliable (retro)synthesis planning algorithms able to design a synthesis route for an input molecule. Retrosynthesis aims to recursively decompose a target compound into simpler molecules\changed{,} forming a synthesis tree. The leaves of the tree are purchasable molecules from which the synthesis process can start\changed{,} and the tree itself is a synthesis recipe. By going bottom-up the tree and performing reactions defined by the tree nodes, one will eventually obtain the target molecule. The construction of such a tree usually consists of two components: a single-step retrosynthesis model that decomposes a molecule 
and a multi-step planning algorithm that guides the recursive decomposition to obtain the full synthesis tree 
\cite{segler2018planning, coley2019robotic, schwaller2020predicting}. 
In this paper, we focus on single-step retrosynthesis, which predicts a reaction that is likely to synthesize a given molecule. 

In practice, many feasible reactions can lead to a given product. Since the success of a synthesis plan depends on factors that may vary over time (e.g.\changed{,} the availability or cost of reactants), the retrosynthesis model should ideally return all possible reactions. In other words, we would like to produce a diverse set of feasible reactions leading to the requested product. However, the available datasets cover only a fraction of feasible reactions, so for many of the included products, a lot of alternative reactions are missing. 
This limitation of current reaction datasets causes two major issues that we address in this paper.

First, the existing retrosynthesis models are not encouraged to explore the space of feasible reactions well. The main contribution of the paper is the development of a RetroGFN model that can explore beyond the dataset and return a diverse set of feasible reactions. RetroGFN is based on the recent GFlowNet framework \cite{bengio2021flow,bengio2023gflownet}\changed{,} which enables exploration of the solution space and sampling from that space with probability proportional to the reward function which we define using an auxiliary reaction feasibility prediction model to guide our RetroGFN training outside the limited reaction dataset. In consequence, our model samples a large number of feasible reactions. It outperforms existing methods on the round-trip accuracy metric while achieving competitive results on the top-k accuracy.

Second, the typical evaluation of retrosynthesis models involves the use of top-k accuracy, which verifies how many top-k reactions returned by the model are included in a given dataset. Our analysis performed on the USPTO-50k test split \cite{schneider2016s,coley2017retrosim,jin2017predicting} reveals that\changed{, on average,} more than 100 feasible reactions returned by the examined retrosynthesis models are ignored by top-k accuracy (see \Cref{fig:ignored_reactions}). Since it is practically impossible to include all possible reactions in a finite dataset, one remedy relies on employing a machine learning model, which reliably assesses the reaction feasibility. This approach is applied in round-trip accuracy \cite{schwaller2020predicting}, a less exploited alternative to the top-k accuracy metric. Round-trip accuracy was shown to correlate more with human judgment than standard top-k accuracy \cite{jaume2023transformer} and recommended as a \changed{complementary} metric in previous works \cite{westerlund2024chemformers,chen2021deep,graph2edit,igashov2023retrobridge}. In this paper, we additionally strengthen the arguments in favor of reporting round-trip accuracy in the papers and demonstrate that replacing top-k accuracy with top-k round-trip accuracy decreases the number of ignored reactions while being robust to non-trivially unfeasible reactions.

To summarize, our contributions are:
\begin{itemize}[leftmargin=15pt,topsep=2pt]
    \setlength{\itemsep}{2pt}
    \setlength{\parskip}{0pt}
    \item We develop RetroGFN: a model based on the GFlowNet framework that generates diverse and feasible reactions. To our knowledge, we are the first to adapt GFlowNets for retrosynthesis. RetroGFN is guided by an auxiliary feasibility model that allows exploration outside the limited reaction dataset (\Cref{sec:retro_gfn}). We make the code publicly available\footnote[1]{\url{https://github.com/gmum/RetroGFN}}.
    \item We benchmark the state-of-the-art single-step retrosynthesis models and show that our RetroGFN outperforms all considered models for $k>3$ on the round-trip accuracy while achieving competitive results on the top-k accuracy (\Cref{sec:benchmarks}). 
    \item We provide empirical arguments for the importance of reporting the round-trip accuracy in the single-step retrosynthesis model evaluation\changed{,} which motivates our RetroGFN (\Cref{sec:ftc_metric}).

\end{itemize}

\section{Related Work}
\label{sec:related_work}
\textbf{Single-step Retrosynthesis.} The single-step retrosynthesis problem is well-known in the drug-discovery community. The methods in this field can be roughly divided into template-based, template-free, and semi-template. Template-based methods use reaction templates (also called rules, see \Cref{fig:template_example}), which explicitly describe the graph-level transformation of molecules that are encountered in the reactions \cite{segler2017neural,coley2017retrosim,dai2019retrosynthesis,baylon2019enhancing}. Templates provide a strong inductive bias as they form a fixed set of possible transformations that the retrosynthesis model can perform. Template-free approaches, on the other hand, do not rely on a template and aim to generate the transformation of the product (the change of the bonds and atoms between reactants and product) \changed{\cite{sacha2021molecule,yan2020retroxpert,graphretro,wang2021retroprime}} or generate the product from scratch \cite{irwin2022chemformer,schwaller2020predicting,zheng2019predicting,mao2021molecular}. The semi-template models typically rely on the rules extracted from the training dataset, but without explicitly building a fixed set of reaction templates \cite{graphretro,graph2edit}. Our RetroGFN is a semi-template model, as it is not limited to a fixed set of templates, but instead composes \changed{them} using pre-defined patterns. The template composition process was inspired by RetroComposer \cite{yan2022retrocomposer} but remains substantially different: we implement the generation process in the GFlowNets framework; we use more general patterns; we parametrize the second phase to be order-invariant; we guarantee the second phase ends with product and reactant patterns that can be mapped; and finally, we map the atoms using a machine learning model (while RetroComposer uses a heuristic).

\textbf{GFlowNets.} GFlowNets \cite{bengio2023gflownet} are a type of generative methods devoted to sampling from high-dimensional distributions. GFlowNets were originally proposed as an alternative to MCMC (offering the benefits of amortization) and reinforcement learning (displaying a mode-seeking behavior, that is\changed{,} the ability to discover multiple diverse modes), and later shown to be equivalent \changed{to} special cases of other generative methods \cite{malkin2022gflownets}. 
The diversity, in particular, is a desired property in multiple scientific discovery tasks \cite{bengio2021flow,jain2023multi,koziarski2024rgfn}.


\section{RetroGFN}
\label{sec:retro_gfn}

RetroGFN is a single-step retrosynthesis model, meaning it predicts a set of molecules that could react to a given target product (see \Cref{fig:template_example} a)). A product is represented as an annotated graph $G=(V, T, E)$, where nodes $V=\{v_1, v_2, ..., v_n\}$ correspond to the molecule's atoms along with associated atom symbols (types) $T$, and edges $E$ are bonds. Additionally, each node and edge \changed{has} an associated vector of features that will be used when embedding a molecule. 

RetroGFN's training is guided by a reaction feasibility model, enabling exploration outside the limited reaction dataset. More details on the guidance can be found in \Cref{sec:training}.

\subsection{Reaction Templates and Patterns}

\begin{wrapfigure}{r}{0.4\textwidth}
    \vspace{-8mm}
  \begin{center}
    \includegraphics[width=0.4\textwidth]{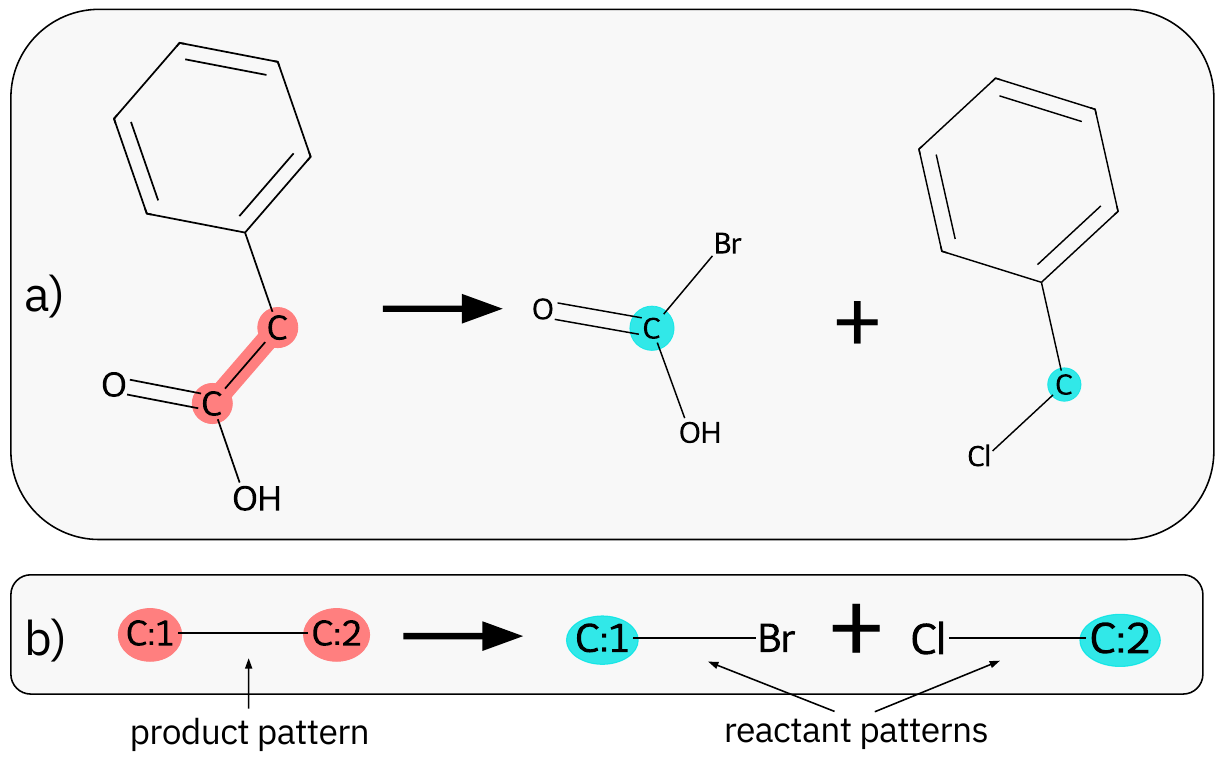}
  \end{center}
  \caption{Illustration of a single-step retrosynthesis (a), and a corresponding reaction template (b). Atoms from a product pattern on the left side of the template are mapped to atoms from reactant patterns on the right side (red C:i is mapped to blue C:i).}
  \label{fig:template_example}
\end{wrapfigure}

Several existing single-step retrosynthesis models, including ours, work on the (backward) reaction templates. 
A reaction template can be seen as a regular expression on graphs (see \Cref{fig:template_example}). It describes the transformation of a product into the reactants and consists of the product pattern 
(left side of the regular expression) and a set of reactants' patterns (right side). The atoms of the product pattern are mapped to \changed{the} atoms of \changed{the} reactants' patterns. Reaction templates provide a strong inductive bias to the model while limiting it to a fixed set of possible transformations. 

However, we extend the covered reaction space by introducing a template composition process inspired by RetroComposer \cite{yan2022retrocomposer}. In this approach, we choose the reaction center where the template is going to be applied and compose a concrete template step by step using the building blocks, called patterns.

We extract the templates from the train split of USPTO-50k, following \cite{chen2021deep}. Each template is then split into product and reactant patterns (see \Cref{fig:template_example} b)). We denote a set of all encountered product patterns $PPS$ and an analogous set of reactant patterns $RPS$. The patterns do not include any molecular regular expression (SMARTS) and can be represented similarly to molecules\changed{,} as annotated graphs. 


\subsection{Generation Process}
\begin{figure*}[t]
\begin{center}
\vskip -0.5in
\centerline{\includegraphics[width=\linewidth]{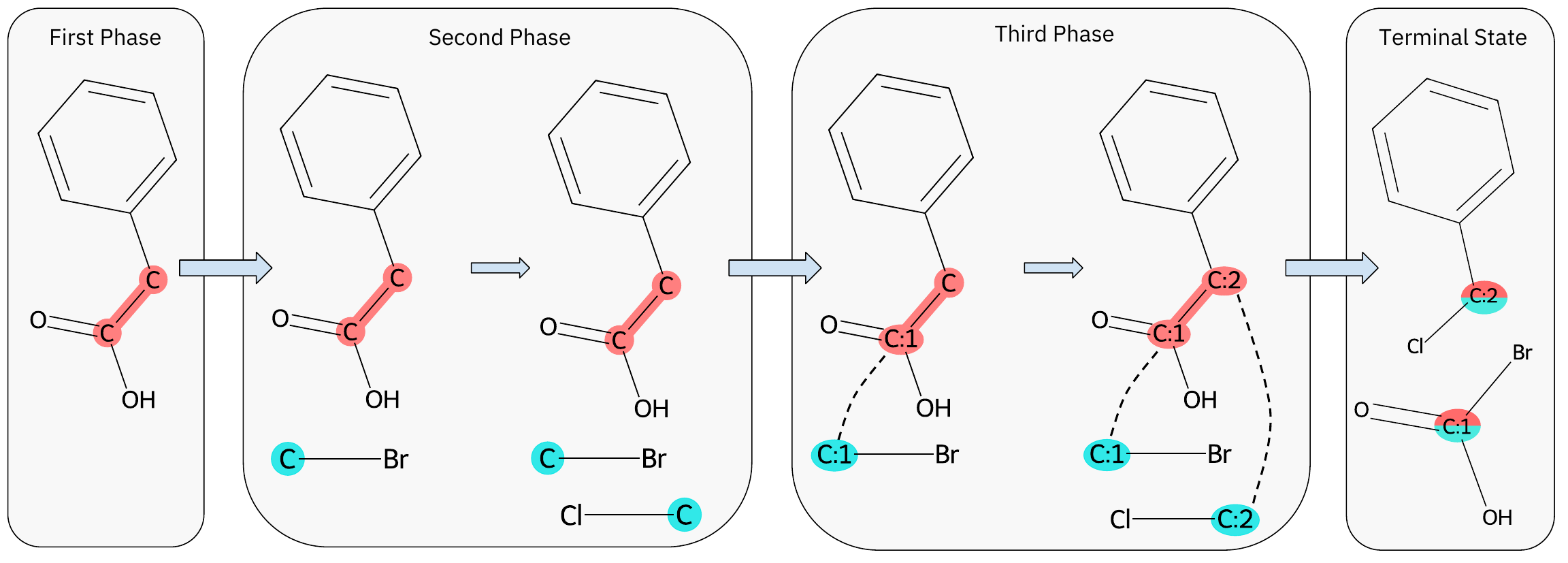}}
\caption{Illustration of the template composition process in RetroGFN for an input product. In the first phase, a product pattern and its concrete match to the atoms of the product \changed{are} chosen. In the second phase, reactant patterns are gathered until all mappable atoms of the product pattern (highlighted red) can be mapped to mappable atoms of the reactant pattern (highlighted blue). In the third phase, the mapping between mappable \changed{products} and reactant patterns is created\changed{,} and the obtained template is applied\changed{,} resulting in the reactants.}
\label{fig:composition_example}
\end{center}
\vskip -0.2in
\end{figure*}

Given a product, our RetroGFN composes an appropriate template in three phases: 
\begin{enumerate}[leftmargin=15pt,topsep=0pt]
    \setlength{\itemsep}{2pt}
    \setlength{\parskip}{0pt}
    \item The first phase determines a reaction center: a product pattern matched with the product.
    \item The second phase gathers the reactant patterns.
    \item The third phase constructs atom mapping between the atoms of the product pattern and the reactants' patterns.
\end{enumerate}
In the end, the obtained template is applied to the given product and results in a final set of reactants. \Cref{fig:composition_example} shows an example of the composition process\changed{,} while a detailed description of each phase can be found further in the section.

The core component of a GFlowNet model is a forward policy $P_F(a | s)$ describing the probability of taking action $a$ in the state $s$. The generation process samples a sequence of states and actions $\tau=(s_1, a_1, ..., s_k, a_k, t)$ called a trajectory, where $t$ is a terminal state. In RetroGFN, an initial state $s_1$ is an input product, the intermediate states $s_i$ correspond to the partially constructed template, and the terminal state $t$ stores a final template along with a result of its application to the product. We group the states into three phases\changed{,} and the specific definition of $P_F(a | s)$ depends on the phase $i$:

\begin{equation*}
    P^i_F(a | s) = \frac{\exp(\score_i(s, a)\alpha)}{\sum_{a' \in A^i(s)}\exp(\score_i(s, a') \alpha)},
\end{equation*}
where $\score_i$ is a phase-specific score function parameterized with a neural network and $A^i(s)$ is a set of possible actions that can be taken from $s$ in the $i$-th phase. The policy is simply a softmax with temperature coefficient $\alpha$ over the scores of all possible actions $A^i(s)$.

Score functions for all the phases share a common Graph Neural Network (GNN) encoder, denoted as $\gnn_1$ that given a product $p=(V, E, T)$, embeds its nodes' features: $\gnn_1(p) \in \R^{n \times d}$, where $n$ is the number of product nodes and $d$ is the embedding size. We overload the notation and let $\gnn_1(v_j)$ denote the embedding of a product node $v_j \in V$. The GNN architecture we use is similar to the one from LocalRetro: a stack of MPNN layers with a single Transformer layer \cite{vaswani2017attention} on top. Details can be found in the \ref{app:retro_gfn_details}.

\textbf{First Phase.} A state $s$ in the first phase is an input product $p$. The action space $A^1(s)$ contains all possible atom matchings of product patterns from $PPS$ to the product $p$. An action $a \in A^1(s)$ contains the matched product pattern $pp \in PPS$ and the matched atom indices $I=\{i_1, ..., i_m\}$. The value of $i_j$ is an index of the product atom matched with $j$-th product pattern atom. To compute the $\score_1(s, a)$, we aggregate the representation of matched \changed{products'} nodes and put them into \changed{a} multi-layer perceptron $\MLP_1: \R^d \rightarrow \R$:
\begin{equation*}
    \score_1(s, a) = \MLP_1\left(\sum_{i \in I} \gnn_1(v_i)\right).
\end{equation*}
After the action is chosen and applied, the generation process transitions directly to the second phase.

\textbf{Second Phase.} 
The second phase iteratively adds reactant patterns to the
\begin{wrapfigure}{r}{0.45\textwidth}
\vskip -10pt
  \begin{center}
    \includegraphics[width=0.45\textwidth]{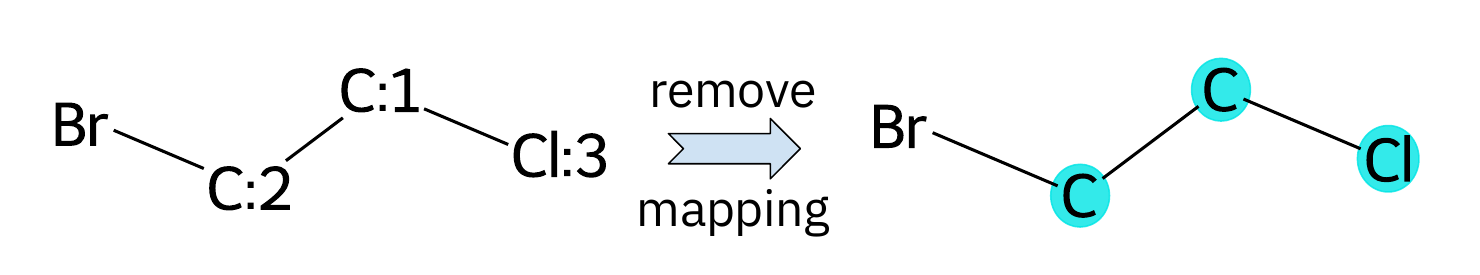}
  \end{center}
    \caption{Illustration of a pattern before (left) and after (right) mapping removal. The mappable atoms of the pattern are colored blue.}
    \label{fig:mappable_example}
    \vskip -10pt
\end{wrapfigure}
 composed template. At the beginning of the phase, the list of reactant patterns is empty. The second phase action $a$ is a reactant pattern $rp_j \in RPS$ that is going to be added to the template. The $\score_2(s, a)$ concatenates the information from the previous phase and the reactant patterns collected so far (denoted as $R$) and feeds it to $\MLP_2: \R^{3d} \rightarrow \R^{|RPS|}$ that predicts the score for all the reactant patterns in $RPS$:
\begin{align*}
    \score_2(s, a)
    =\MLP_2\biggl(\sum_{i \in I} \gnn_1(v_i) \; | \;
    \E_{PPS}(pp) \; | \sum_{rp \in R}\E_{RPS}(rp)\biggr)_j.
\end{align*}
Here we select the $j$th score returned by the $\MLP_2$ as it corresponds to $rp_j$ reactant pattern from the action. Index embedding $e=\E_{A}(a)$ is a function that looks up the index of the element $a$ in the set $A$ and assigns the index a learnable embedding $e \in \R^d$ (e.g. $\E_{PPS}$ assign a unique learnable embedding to every $pp \in PPS$).
At the end of this phase, we want to be sure that every atom from the product pattern can be mapped to some atom of the reactant pattern. Originally, each pattern had some atom mapping in the template it comes from (see \Cref{fig:template_example}). Although those explicit mappings are inadequate in the novel-composed template, we can leverage the knowledge that an atom was originally mapped. For every pattern, we construct a set of mappable atoms that consists of the pattern's atoms that were mapped in the original template (see \Cref{fig:mappable_example}). The composed template is allowed to map only the mappable atoms. We ensure that all mappable atoms in the composed template can be mapped by properly restricting the action space $A^2(s)$.

\textbf{Third Phase.} The third phase creates a mapping between atoms of product and reactant patterns. An action $a$ is an atom mapping $(j, k, l) \in M$ that links the $j$-th node from the product pattern $pp$ with the $l$-th mappable node of the $k$-th reactant pattern from the list of reactant patterns $R$.  The $\score_3(s, a)$ is given with the formula:
\begin{align*}
\score_3(s, a) = \MLP_3(\gnn_1(v_{i_j}) \; |\; \gnn_2(v_{kl})),
\end{align*}
where $v_{i_j}$ is a product node matched with the $j$-th node of the product pattern, and $v_{kl}$ is the $l$-th node of the $k$-th reactant pattern from $R$. To embed the reactant pattern nodes, we introduce a GNN $\gnn_2$ with the same architecture as $\gnn_1$. The action space $A^3(s)$ contains all possible atom mappings. We call an atom mapping between two nodes possible when the atom symbols of the nodes are the same and neither of the nodes \changed{has been} previously mapped. The third phase ends when every node from the product pattern is mapped, resulting in a template that can be applied to the reaction center chosen in the first phase. The obtained reaction forms the terminal state $t$.

\subsection{Training}
\label{sec:training}
We trained our RetroGFN with a modified version of Trajectory Balance Objective from \cite{malkin2022trajectory}, which for a trajectory $\tau = (s_1, a_1, s_2, a_2, ..., s_k, a_k, t)$ is given with the formula:
\begin{equation*}
    \mathcal{L}(\tau) = \left(  \log \frac{F(s_1)\prod_{i=1}^k P_F(a_i | s_i)}{R(t)P_B(a_{k} | t)\prod_{i=2}^kP_B(a_{i-1} | s_i)} \right)^2.
\end{equation*}
The main difference from the original formulation comes from the fact that our RetroGFN is conditioned on the product from the initial state $s_1$. Therefore, for every initial state, we estimate the incoming flow separately using $F(s_1)$ function which is essentially an index embedding $F(s)=E_{P}(s) \in \R$ that looks up the set of training products $P$ and returns a learnable scalar (note that we only evaluate $F(s)$ during training). As a backward policy $P_B(a | s)$, we use a uniform distribution over the possible actions that could lead to state $s$. The reward is an exponential reward of the form $R(x) = \exp(\beta f(x))$ where $f$ is a proxy for the desired property. 

\paragraph{Reaction Feasiblity Guidance}. In RetroGFN, we use a reaction feasibility model as a proxy $f$. This way, we allow for exploration outside the limited reaction dataset as RetroGFN can assess the reaction feasibility via proxy $f$. The feasibility proxy can be a machine learning model that predicts the feasibility of a reaction or an indicator of whether the forward reaction prediction model was able to backtranslate the reaction $x$. In the main part of the paper, we evaluate the latter, while experiments with the former can be found in the \Cref{app:rfm}. Importantly, the forward model used during training was trained only on the training set of USPTO-50k, making it distinct from the model used in the round-trip evaluation.

\subsection{Inference}
\label{sec:inference}
During inference, the retrosynthesis model is given a product and requested to output at most $N$ reactions sorted from the most to least promising. RetroGFN samples the reactions using the trained forward policy $P_F(a | s)$ and orders them with the estimated probability. The probability of a reaction represented by a terminal state $t$ is estimated by summing the probabilities of all sampled trajectories that end with $t$: 
\begin{equation}
    p(t) = \sum_{\tau: t \in \tau} \prod_{(s, a) \in \tau} P_F(a | s).
\end{equation}
To increase the accuracy of the estimation, we sample $K \cdot N$ trajectories. We leave the exploration of other estimation methods for future work. The details on the architecture and hyperparameters of both training and inference can be found in the \ref{app:retro_gfn_details}.

\begin{table}[t]
    \centering
    \caption{Top-k round-trip accuracy results on USPTO-50k along with the standard deviation. The models are grouped into template-free, template-based, and semi-template models, respectively. The best results in every column are \textbf{bolded}. We observe that for $k>1$ our RetroGFN consistently outperforms other methods.}
    \label{tab:ftc_50k}
    \begin{tabular}{@{}lcccccc@{}}
    \toprule
        model & top-1 & top-3 & top-5 & top-10 & top-20 & top-50 \\ 
        \midrule
        MEGAN & 88.2 & 81.4 & 77.0 & 68.8 & 58.3 & 41.8 \\ 
        RootAligned & \textbf{94.5} & 85.3 & 79.0 & 67.8 & 53.3 & 27.1 \\ 
        Chemformer & 92.6 & 58.9 & 42.0 & 24.2 & 12.9 & 5.3 \\ 
        \midrule
        GLN & 93.4 & 87.6 & 84.2 & 77.8 & 67.5 & 47.5 \\ 
        MHNreact & 91.9 & 85.6 & 81.0 & 72.1 & 60.1 & 38.5 \\ 
        LocalRetro & 94.2 & 88.3 & 85.1 & 79.6 & 71.2 & 54.3 \\ 
        RetroKNN & 94.0 & 87.8 & 84.3 & 78.7 & 70.0 & 53.1 \\ 
        \midrule
        Graph2Edits & 92.6 & 82.8 & 75.9 & 61.5 & 42.8 & 21.3 \\ 
        GraphRetro & 93.1 & 70.0 & 60.4 & 45.5 & 30.2 & 14.6 \\ 
        RetroGFN & 92.8 & \textbf{88.8} & \textbf{86.1} & \textbf{81.4} & \textbf{74.9} & \textbf{63.6} \\   \bottomrule
    \end{tabular}
\end{table}

\begin{table}[t]
    \centering
    \caption{
        Top-k accuracy and mean reciprocal rank (mrr) results on USPTO-50k. The models are grouped into template-free, template-based, and semi-template models, respectively. The numbers in columns denote $k$ values. The best results in every column are \textbf{bolded}. We observe that for $k>3$ our RetroGFN achieves competitive results.
    }
    \label{tab:standard_50k}
    \begin{tabular}{@{}lccccccc@{}}
    \toprule
        method & mrr & top-1 & top-3 & top-5 & top-10 & top-20 & top-50 \\ 
        \midrule
        MEGAN & 0.6231 & 48.7 & 72.4 & 79.5 & 86.8 & 91.0 & 93.5 \\ 
        RootAligned & \textbf{0.6886} & \textbf{56.0} & \textbf{79.1} & \textbf{86.1} & 91.0 & 93.3 & 94.2 \\ 
        Chemformer & 0.6312 & 55.0 & 70.9 & 73.7 & 75.4 & 75.9 & 76.0 \\ 
        GLN & 0.6509 & 52.4 & 74.6 & 81.2 & 88.0 & 91.8 & 93.1 \\ 
        \midrule
        MHNreact & 0.6350 & 50.8 & 72.7 & 79.6 & 86.3 & 90.0 & 92.3 \\ 
        LocalRetro & 0.6587 & 52.0 & 76.6 & 84.6 & 91.1 & \textbf{94.9} & \textbf{96.7} \\ 
        RetroKNN & 0.6837 & 55.7 & 77.9 & 85.5 & \textbf{91.7} & \textbf{94.7} & \textbf{96.5} \\ 
        \midrule
        Graph2Edits & 0.6684 & 54.6 & 76.6 & 82.8 & 88.7 & 91.1 & 91.7 \\ 
        GraphRetro & 0.6087 & 53.7 & 67.0 & 70.6 & 73.4 & 74.0 & 74.2 \\ 
        RetroGFN & 0.6308 & 49.2 & 73.3 & 81.1 & 88.0 & 92.2 & 95.3 \\ \bottomrule
    \end{tabular}
    \vskip -10pt
\end{table}

\begin{table*}[t]
    \centering
    \caption{Top-k round-trip accuracy on USPTO-MIT along with the standard deviation. The models are grouped into template-free, template-based, and semi-template models, respectively. The best results in every column are \textbf{bolded}. We observe that for $k>3$ our RetroGFN consistently outperforms other methods.}
    \label{tab:ftc_mit}
    \begin{tabular}{@{}lcccccc@{}}
    \toprule
        model & top-1 & top-3 & top-5 & top-10 & top-20 & top-50 \\ 
        \midrule
        MEGAN & 83.2 & 75.8 & 71.3 & 63.5 & 54.1 & 39.4 \\ 
        RootAligned & \textbf{85.9} & 76.6 & 70.6 & 60.7 & 48.1 & 24.8 \\ 
        Chemformer & 82.8 & 53.5 & 38.9 & 22.8 & 12.2 & 5.0 \\ 
        \midrule
        GLN & 84.2 & 77.5 & 73.6 & 66.9 & 57.1 & 39.4 \\ 
        MHNreact & 83.3 & 76.5 & 72.2 & 64.0 & 52.6 & 32.7 \\ 
        LocalRetro & 85.3 & \textbf{79.2} & \textbf{76.0} & 70.4 & 62.8 & 48.0 \\ 
        RetroKNN & 85.3 & 77.5 & 72.9 & 62.5 & 42.9 & 18.6 \\ 
        \midrule
        Graph2Edits & 83.9 & 74.3 & 67.9 & 55.5 & 39.4 & 20.3 \\ 
        GraphRetro & 84.3 & 63.0 & 53.6 & 39.4 & 25.6 & 12.3 \\ 
        RetroGFN & 83.9 & \textbf{78.8} & \textbf{76.0} & \textbf{71.4} & \textbf{65.8} & \textbf{56.2} \\  \bottomrule
    \end{tabular}
\end{table*}

\begin{table}[t]
    \centering
    \caption{
        Top-k accuracy and mean reciprocal rank (mrr) results on USPTO-MIT. The models are grouped into template-free, template-based, and semi-template models, respectively. The best results in every column are \textbf{bolded}. We observe that our RetroGFN achieves competitive results.
    }
    \label{tab:standard_mit}
    \begin{tabular}{@{}lccccccc@{}}
    \toprule
        method & mrr & top-1 & top-3 & top-5 & top-10 & top-20 & top-50 \\ 
        \midrule
        MEGAN & 0.4647 & 37.0 & 53.4 & 58.8 & 63.7 & 66.8 & 68.9 \\ 
        RootAligned & \textbf{0.4960} & \textbf{40.2} & \textbf{56.8} & \textbf{61.7} & \textbf{66.1} & \textbf{68.5} & 69.5 \\ 
        Chemformer & 0.4521 & 39.5 & 50.5 & 52.6 & 53.8 & 54.1 & 54.2 \\ 
        \midrule
        GLN & 0.4625 & 37.2 & 52.9 & 57.7 & 62.7 & 65.2 & 66.3 \\ 
        MHNreact & 0.4610 & 37.2 & 52.7 & 57.6 & 62.1 & 64.7 & 66.2 \\ 
        LocalRetro & 0.4720 & 36.7 & 55.1 & 60.8 & \textbf{66.0} & \textbf{68.7} & \textbf{70.5} \\ 
        RetroKNN & 0.4572 & 35.6 & 53.3 & 59.1 & 64.3 & 66.2 & 66.4 \\ 
        \midrule
        Graph2Edits & 0.4795 & 38.9 & 55.2 & 60.1 & 64.0 & 65.6 & 66.1 \\ 
        GraphRetro & 0.4461 & 39.1 & 49.2 & 51.8 & 53.7 & 54.3 & 54.6 \\ 
        RetroGFN & 0.4590 & 35.4 & 53.5 & 59.3 & 64.7 & 67.8 & \textbf{70.0} \\  \bottomrule
    \end{tabular}
\end{table}

\section{Experiments}
\label{sec:benchmarks}
This section describes the benchmark methodology and results of our RetroGFN models compared to the current state-of-the-art. Tables \ref{tab:ftc_50k}, \ref{tab:standard_50k}, \ref{tab:ftc_mit} and \ref{tab:standard_mit} show that our RetroGFN outperforms all considered models on round-trip accuracy while achieving competitive results on the top-k accuracy.

\subsection{Setup}
\label{sec:setup}
\textbf{Datasets.} We compared the considered methods on two datasets: USPTO-50k, a default choice for benchmarking retrosynthesis models, and USPTO-MIT, which we use as a generalization benchmark for models trained on USPTO-50k. We used commonly used splits for both datasets \cite{coley2017retrosim,jin2017predicting}. We refined the USPTO-MIT to ensure there is no overlap between it and the USPTO-50k train split.

\textbf{Retrosynthesis Models.} We compared our RetroGFN to well-known and recent state-of-the-art retrosynthesis models; template-free: MEGAN~\cite{sacha2021molecule}, RootAligned~\cite{zhong2022root} and Chemformer~\cite{irwin2022chemformer}; template-based: GLN~\cite{dai2019retrosynthesis}, , MHNreact~\cite{seidl2021modern}, LocalRetro~\cite{chen2021deep}, RetroKNN~\cite{xie2023retroknn}; semi-template-based: Graph2Edits \cite{graph2edit}, GraphRetro \cite{graphretro}. We used the wrappers of the original implementations and checkpoints from the Syntheseus repository\footnote{\url{https://github.com/microsoft/syntheseus}}.
We used the evaluation procedure from Syntheseus that queries the model for 100 reactions, removes the duplicates, and truncates the list of reactions for every product to be no larger than 50. The same output was used both to calculate standard and round-trip metrics. All models were trained using the same train/validation/test splits.

\textbf{Forward Model.}
A forward (reaction prediction) model takes a set of reactants as an input and outputs a set of possible products. As a backbone, we used a pre-trained Chemformer model from \cite{irwin2022chemformer}. We fine-tuned two forward models: Chemformer-Eval\changed{,} which was used to estimate the reaction feasibility in the round-trip accuracy (see \cref{sec:ftc_metric})\changed{,} and Chemformer-Train\changed{,} which guided RetroGFN during the training (see \cref{sec:training}). Chemformer-Train was fine-tuned on the train split of USPTO-50k, while Chemformer-Eval used both the train and test split of USPTO-50k.

\subsection{Results on USPTO-50k}
The top-k round-trip accuracy results for the USPTO-50k dataset can be found in \Cref{tab:ftc_50k}. Note that the forward model used during the training of RetroGFN was trained on a different data split than the one used for evaluation. We observe that for $k>1$ RetroGFN consistently outperforms all the models. The absolute and relative advantage of RetroGFN over the second-best model on top-k round-trip increases with $k$, which indicates two things. First, the model can return many diverse and feasible reactions while other models struggle to do that for $k>10$ (e.g.\changed{,} Chemformer). Second, while RetroGFN returns a lot of high-quality candidates, it does not rank them optimally. Therefore, its results on metrics that are sensitive to the quality of ranking, namely metrics with low $k$, are impoverished. We believe that improving the RetroGFN's inference described in \Cref{sec:inference} may be a remedy for its performance with low $k$, but we leave it for future work.

In \Cref{tab:standard_50k}, we can find standard top-k accuracy results. Our method performs competitively with state-of-the-art single-step retrosynthesis models, especially for larger values of $k$\changed{,} which are arguably more important for retrosynthesis search than $k=1$.

The good results of RetroGFN on standard metrics and its performance on round-trip accuracy
evidence that one can improve the results on round-trip accuracy without sacrificing the performance
on standard metrics, especially for larger values of k. Interestingly, the Pearson correlation between
top-k accuracy and top-k round-trip for \changed{k=1} seems relatively high (corr=0.64, p-value=0.08) and
increases with k(corr=0.85 and p-value=0.006 for \changed{k=50}), indicating that there may be a space for
jointly maximizing both metrics.

\subsection{Generalization Results on USPTO-MIT}
We evaluated the models trained on USPTO-50k further on the USPTO-MIT dataset to assess their generalization properties (\Cref{tab:ftc_mit} and \ref{tab:standard_mit}). The evaluation of both standard and round-trip accuracy metrics echoes the results of USPTO-50k: RootAligned is the best on top-k accuracy, while our model achieves SOTA results on round-trip metrics. As in the USPTO-50k case, the absolute and relative advantage of RetroGFN over the second-best model on top-k round-trip increases with k. 

\subsection{Diversity}
\begin{figure*}[t]
\begin{center}
\label{fig:scaffolds}
\centerline{\includegraphics[width=\linewidth]{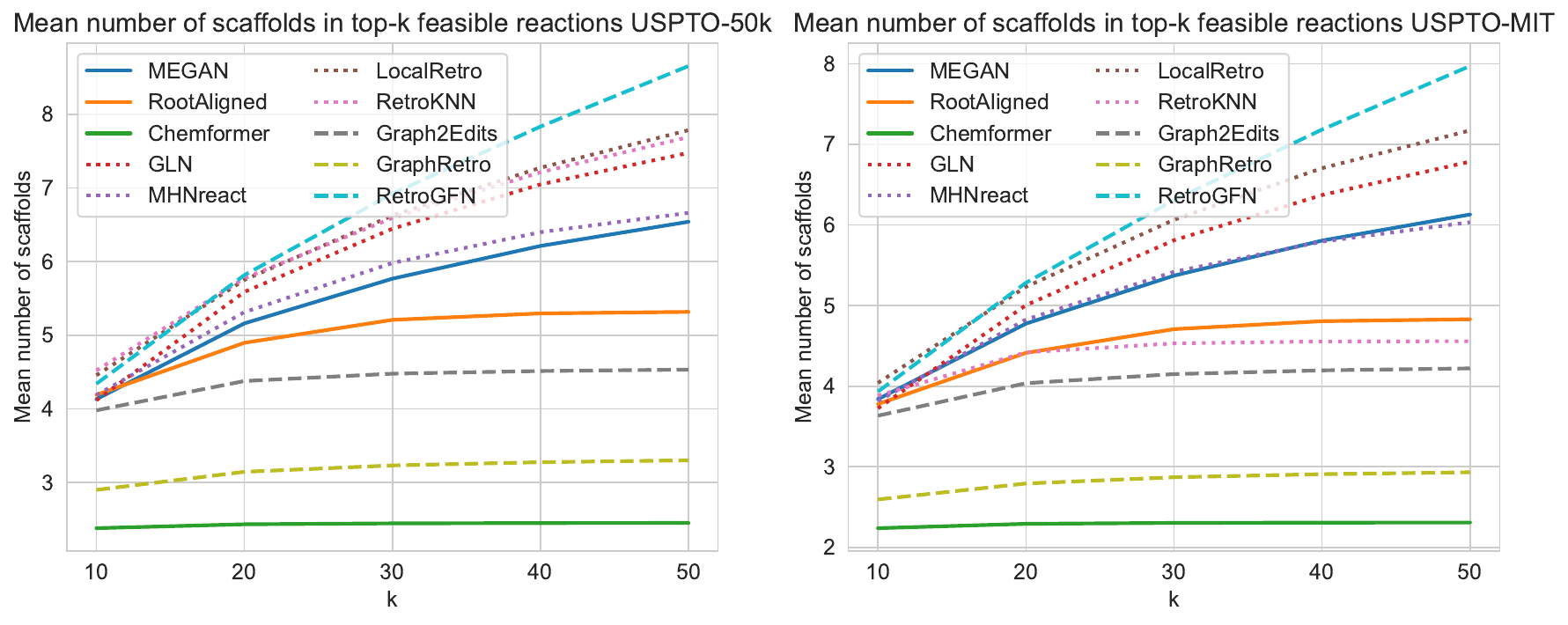}}
\caption{A plot showing the diversity of the reactions proposed by various single-step retrosynthesis models. It shows \changed{the} mean number of distinct molecular scaffolds observed in the top-k returned reactions that were predicted as feasible by a forward reaction prediction \changed{model} (fine-tuned on USPTO-50k and USPTO-MIT\changed{,} respectively). We observe that RetroGFN is able to return visibly more diverse reactions than other models.}
\vskip -40pt
\end{center}
\end{figure*}

We argue that the diversity of the reactants proposed by the model is captured by round-trip accuracy (\Cref{tab:ftc_50k} and \ref{tab:ftc_mit}), which counts the number of unique reactions that were predicted as feasible by some auxiliary model. Such a uniqueness-based notion of diversity is used in the GFlowNets literature \cite{bengio2021flow}, and we believe \changed{it} is aligned with the retrosynthesis problem, where the reactions proposed for a single product are inherently similar (they share a significant number of atoms). However, to further \changed{assess} the diversity of the models, we additionally reported the number of unique molecular scaffolds discovered among feasible reactions. \Cref{fig:scaffolds} shows that RetroGFN returns diverse sets of reactions, especially for larger k values.

\subsection{Inference Time}
\label{sec:inference_time}
As inference time is an important factor for retrosynthesis prediction, we report it in \Cref{tab:inference_time} for all the considered models for USPTO-50k and USPTO-MIT test sets. We observe a large variety of inference times. For USPTO-50k\changed{,} it ranges from $0.39s$ per molecule for LocalRetro to $15.11s$ for Graph2Edits. Our RetroGFN has a moderate inference time: it ranks $7/10$ on USPTO-50k and $5/10$ on USPTO-MIT. The inference time \changed{can be} reduced by optimizing the inference formulas from \Cref{sec:inference}\changed{,} which we leave for future work. We obtain almost 2x faster inference on USPTO-MIT because we set different inference parameters for this dataset (details in \ref{app:retro_gfn_details}). All times are reported using \changed{the} Syntheseus repository \cite{syntheseus} and V100 Nvidia GPUs.

\begin{table}[!ht]
    \centering
    \label{tab:inference_time}
    \caption{Table with average inference time per molecule in seconds for USPTO-50k and USPTO-MIT test sets.}
    \begin{tabular}{lcc}
    \toprule
        method & USPTO-50k & USPTO-MIT \\ 
        \midrule
        MEGAN & 0.74 & 0.76 \\ 
        RootAligned & 1.55 & 4.49 \\ 
        Chemformer & 6.35 & 5.79 \\ 
        GLN & 0.88 & 0.71 \\ 
        MHNreact & 1.31 & 8.67 \\ 
        LocalRetro & 0.39 & 0.43 \\ 
        RetroKNN & \textbf{0.24} & \textbf{0.25} \\ 
        Grap2Edits & 15.11 & 12.88 \\ 
        GraphRetro & 8.52 & 4.3 \\ 
        RetroGFN & 3.7 & 1.99 \\ 
        \bottomrule
    \end{tabular}
\end{table}

\subsection{Multi-step Retrosynthesis}
\label{sec:multistep}
\begin{figure}
    \label{fig:multistep_res}
    \centering
    \renewcommand{\thesubfigure}{\alph{subfigure}}
    \begin{tabular}{cc}  
        \vspace{-5mm}
        \subfloat{\includegraphics[width=0.45\textwidth]{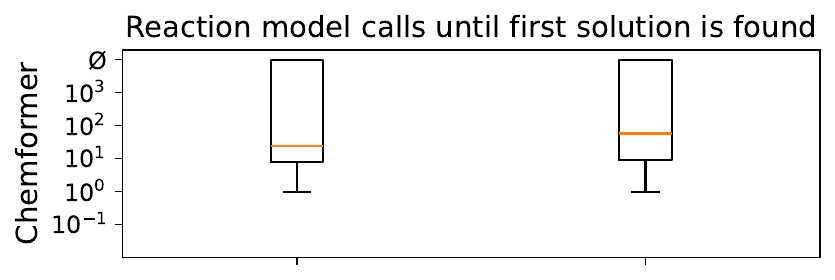}} &
        \subfloat{\includegraphics[width=0.45\textwidth]{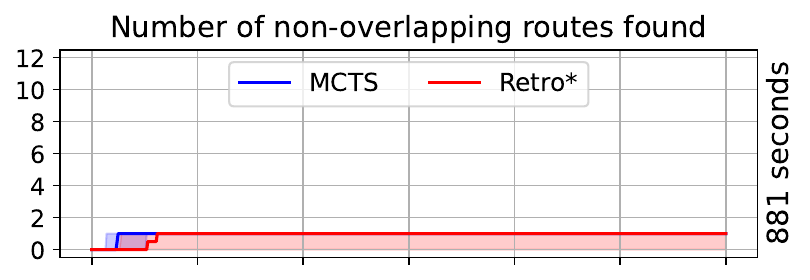}} \\
        \vspace{-5mm}
        \subfloat{\includegraphics[width=0.45\textwidth]{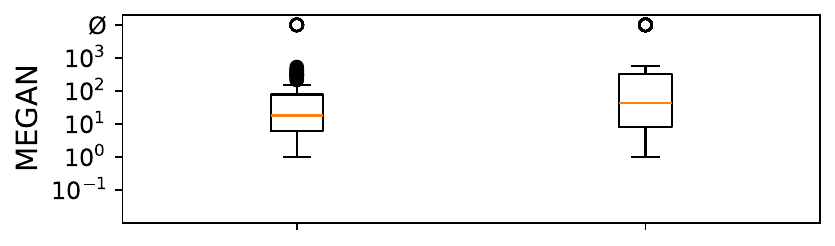}} &
        \subfloat{\includegraphics[width=0.45\textwidth]{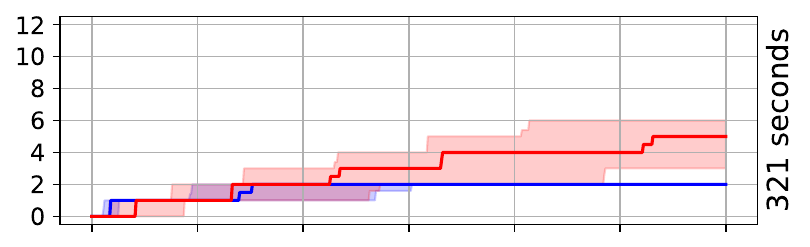}} \\
        \vspace{-5mm}
        \subfloat{\includegraphics[width=0.45\textwidth]{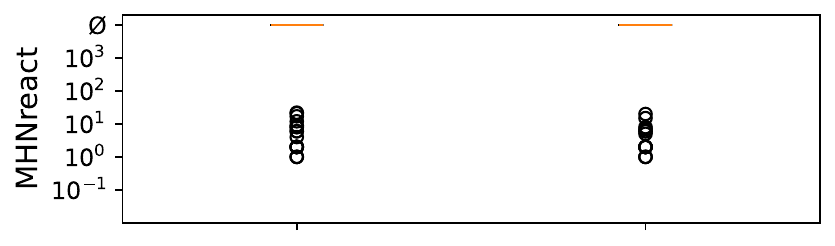}} &
        \subfloat{\includegraphics[width=0.45\textwidth]{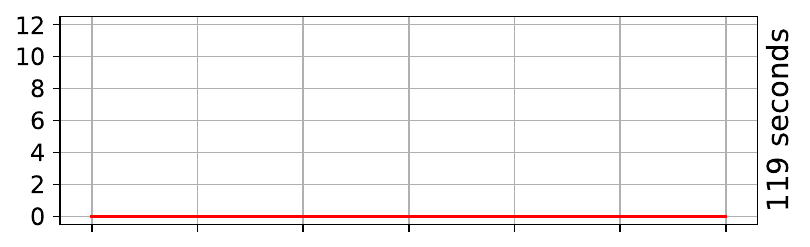}} \\
        \vspace{-5mm}
        \subfloat{\includegraphics[width=0.45\textwidth]{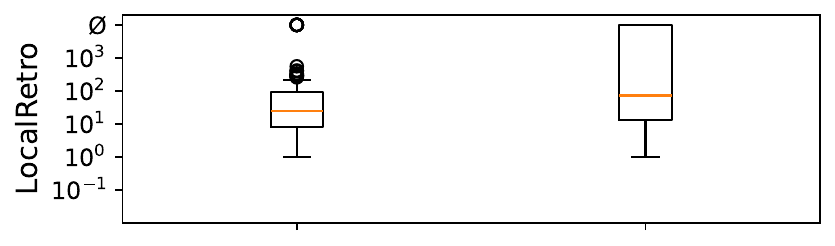}} &
        \subfloat{\includegraphics[width=0.45\textwidth]{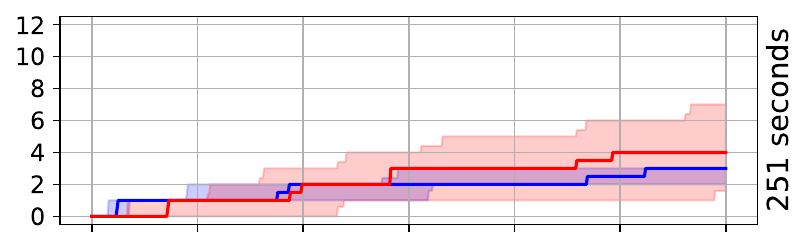}} \\
        \vspace{-5mm}
        \subfloat{\includegraphics[width=0.45\textwidth]{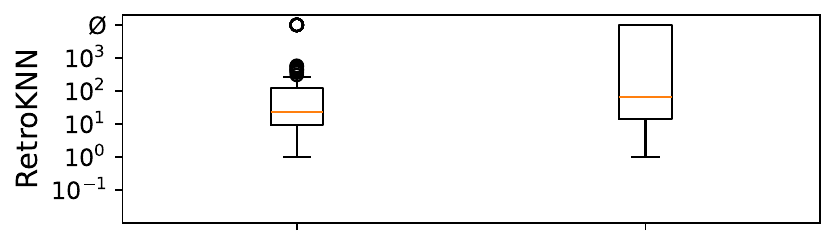}} &
        \subfloat{\includegraphics[width=0.45\textwidth]{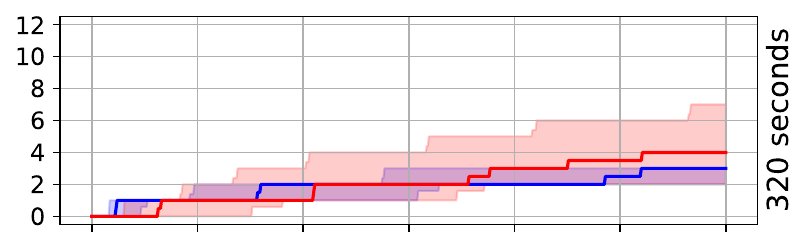}} \\
        \vspace{-5mm}
        \subfloat{\includegraphics[width=0.45\textwidth]{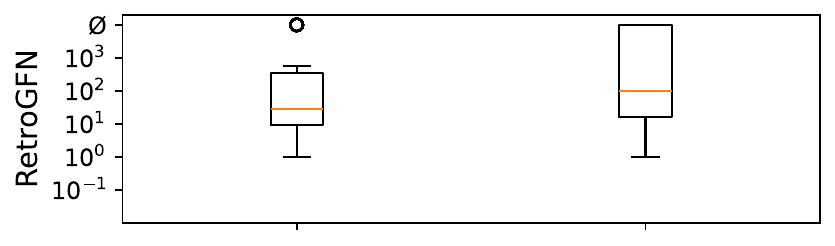}} &
        \subfloat{\includegraphics[width=0.45\textwidth]{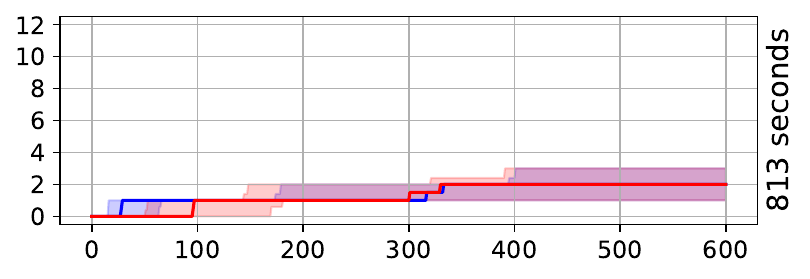}} \\
        \subfloat{\includegraphics[width=0.45\textwidth]{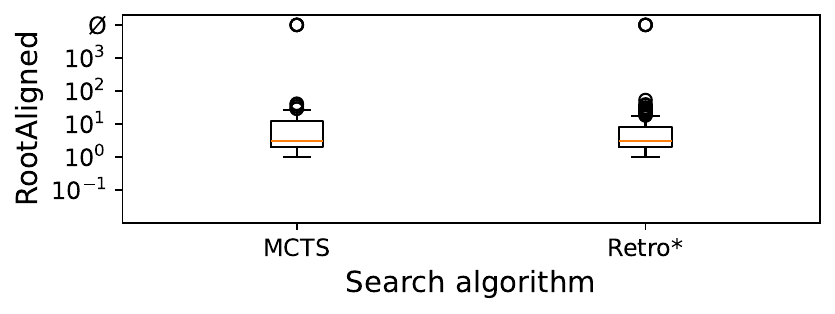}} &
        \subfloat{\includegraphics[width=0.45\textwidth]{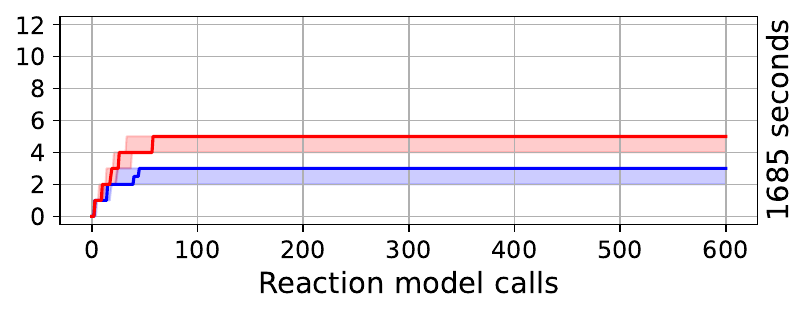}} \\

    \end{tabular}
    \caption{Multi-step search results on the Retro* Hard target set with different single-step models. \textbf{Left}: The number of calls until the first solution was found (or $\emptyset$ if a molecule was not solved). The orange line represents the median, the box represents the 25th and 75th percentile, the whiskers represent the 5th and 95th percentile, and points outside this range are shown as dots. \textbf{Right}: Approximate number of non-overlapping routes present in the search graph (tracked over the number of single-step model calls). The solid line represents the median, shaded area shows the 40th and 60th \changed{percentiles}. On the right-hand side, we note the average time of solving the molecule.}
\end{figure}

We evaluated single-step synthesis models in the multi-step search scenario using \changed{the} Syntheseus repository \cite{syntheseus}. We run each model on Monte Carlo Tree Search (MCTS) and Retro* \cite{chen20retrostar} multi-step algorithms on 190 hard molecules from \cite{chen20retrostar}. Each model is allowed to be called at most 600 times per target molecule and \changed{has the search time limited} to $30$ minutes. \Cref{fig:multistep_res} shows the average number of calls required to solve a molecule and the number of non-overlapping ground-truth synthesis routes. Our RetroGFN achieves moderate results, being outperformed by models with higher top-k accuracy, e.g.\changed{,} LocalRetro, RootAligned or RetroKNN. According to \cite{hartog2024investigations}, the diversity of the single-step candidates and the quality of their ranking are important factors of multi-step search efficiency. The results from \Cref{tab:ftc_50k}, \ref{tab:standard_50k} and \Cref{fig:scaffolds} suggest that RetroGFN outputs diverse candidates (great results for $k>10$), but struggles to rank them optimally (worse results for $k<3$) which diminishes its performance in the multi-step setting. We believe that improving the RetroGFN's inference (\Cref{sec:inference}) is an important future direction to enhance its real-world performance. 

\subsection{Leveraging the Forward Model}
In \ref{sec:ablations}, we study a simple model-agnostic way of leveraging the Chemformer-Train to maximize the results of the round-trip accuracy metric. While this approach significantly improves the round-trip accuracy results, it drastically decreases the standard top-k accuracy, especially for larger values of $k$. We leave the development of other methods of incorporating the Chemformer-Train model into the training pipeline for future work.

\section{Importance of Round-Trip Accuracy}
\label{sec:ftc_metric}

In this section, we set up the single-step retrosynthesis problem, discuss the limitations of the widely used top-k accuracy metric, and argue for the relevance of the round-trip accuracy.

\subsection{Single-Step Retrosynthesis}

Single-step retrosynthesis is focused on predicting reactions that could lead to the given product (see \Cref{fig:template_example}(a)). The retrosynthesis model is evaluated with a reaction dataset $D=\{(R_1, p_1), ..., (R_n, p_n)\}$ containing reaction tuples where $p_i$ denotes a product and $R_i$ is a set of reactants that can synthesize the product $p_i$. During inference, the model is requested to return at most $k$ reactions for every product from the dataset, which are expected to be sorted from the most to the least probable.


\begin{figure*}
\centering
\vskip -30pt
\begin{minipage}{.45\textwidth}
  \centering
  \includegraphics[width=\linewidth]{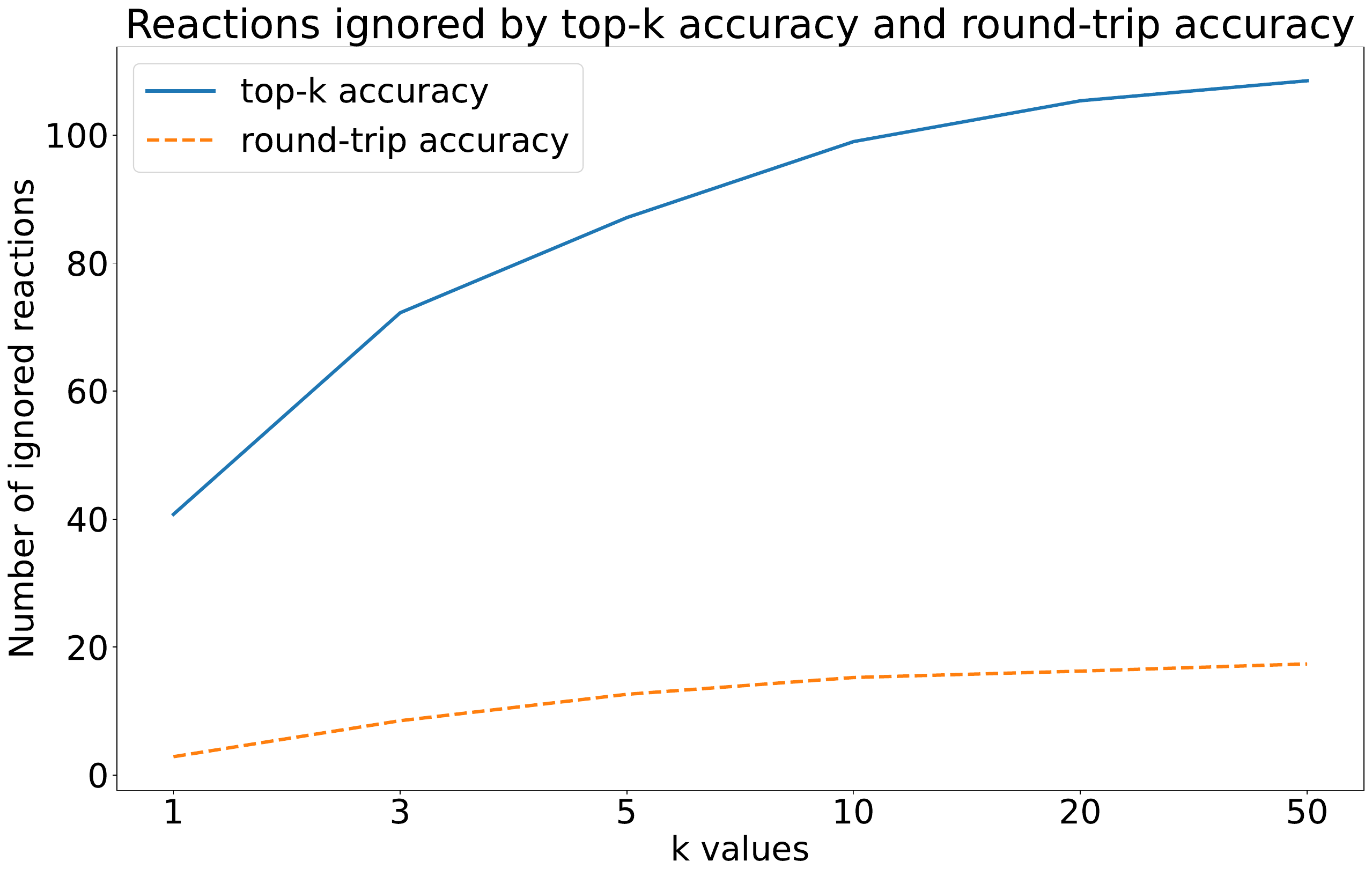}
  \caption{The average number of feasible reactions returned by a retrosynthesis model within its top-k predictions but ignored by top-k accuracy and top-k round-trip accuracy metrics on the USPTO-50k test split. The results are averaged over all models considered in this paper. Ground-truth reaction feasibility was assessed by checking for their presence in USPTO-MIT. We observe that round-trip accuracy treats almost all reactions ignored by top-k accuracy as feasible.}
  \label{fig:ignored_reactions}
\end{minipage}%
\qquad
\begin{minipage}{.45\textwidth}
  \centering
  \includegraphics[width=\linewidth]{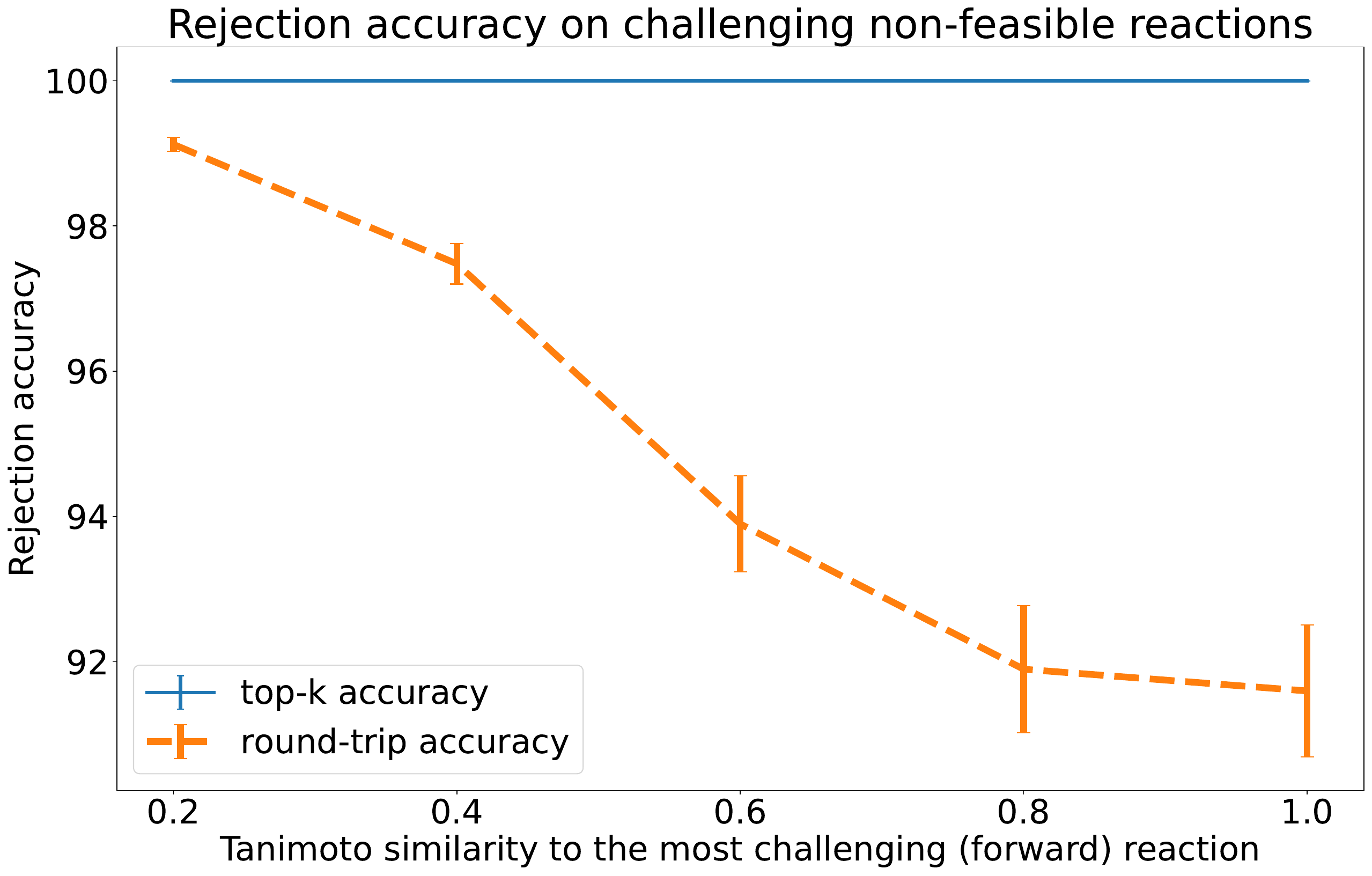}
  \caption{The accuracy of rejecting challenging non-feasible reactions. By definition, the top-k accuracy rejects all non-feasible reactions, while round-trip accuracy treats some portion of them as feasible\changed{,} as it assesses the feasibility with a machine learning model. We observe that even for the most tricky non-feasible reactions, the round-trip accuracy will likely not consider them feasible. Error bars are binomial proportion confidence intervals for $p=0.05$.}
  \label{fig:rejection_accuracy}
\end{minipage}
\end{figure*}





\subsection{Definition of Top-k Accuracy}
Top-k accuracy is one of the most widely used metrics in retrosynthesis. To calculate it for the entire dataset, we first compute the support function $\mathrm{F_{ACC}}$ for every product $p$, which informs whether the ground-truth reaction was found in \changed{the} top $k$ results returned by the model $g$: 
\begin{equation}
    \mathrm{F_{ACC}}(g, p, k) = \mathbb{1}[\exists_{i \leq k}(g(p)_i, p) \in D],
\end{equation}
where $g(p)_i$ is the $i$-th set of reactants proposed by the model for product $p$. Top-k accuracy denotes the portion of ground-truth reactions that were retrieved by the model and can be written as $\mathrm{ACC}(g, k) = \frac{1}{n}\sum_{i=1}^n\mathrm{F_{ACC}}(g, p_i, k)$.

\subsection{Definition of Round-Trip Accuracy}
\label{sec:round_trip_definition}
The top-k round-trip metric uses the wider notion of feasibility than top-k accuracy.
For a single product, the top-k round-trip accuracy value denotes the percentage of feasible reactions among \changed{the} top $k$ reactions returned by the machine learning model. The feasibility is estimated with a forward reaction prediction model $f$\changed{,} which Chemformer \cite{irwin2022chemformer} fine-tuned on USPTO-50k train and test sets (see \Cref{sec:setup}). The exact formula for top-k round-trip accuracy calculated on a product $p$ and retrosynthesis model $g$ is given by:

\begin{equation}
    \mathrm{F_{Round}}(g, p, k) = \frac{1}{k} \sum_{i=1}^k \mathbb{1}[p \in F(g(p)_i)],
\end{equation}
where $f(g(p)_i)$ is the set of products predicted by the forward model $f$ for a set of reactants $g(p)_i$ (we use beam size $=1$). In other words, the metric measures how many reactions proposed by a backward model $g$ can be back-translated by a forward model $f$. We report the top-k round-trip accuracy for the entire dataset $D$, which can be written as $Round(g, k) = \frac{1}{n}\sum_{i}^n\mathrm{F_{Round}}(g, p_i, k)$. Therefore, round-trip accuracy assesses both the diversity and feasibility of the returned reactions.

\subsection{Top-k Accuracy vs Round-Trip Accuracy}
Top-k accuracy works under the assumption that all sensible reactions for a given product are contained in the dataset. However, since there are often many different ways to make a product, it would be too expensive to try all of them. Therefore, real datasets are highly incomplete. In particular, it turns out that this assumption is not true for the USPTO-50k dataset \cite{lowe2012extraction,schneider2016s}\changed{,} which is the most widely used benchmark in the retrosynthesis community. To showcase that, we gathered all reactions returned by any of the considered retrosynthesis models (see \Cref{sec:benchmarks}) that are not included in USPTO-50k. Among 8409 reactions ranked top-1 by any considered model, 76 of them can be found in the USPTO-MIT dataset \cite{jin2017predicting}. While top-k accuracy ignores these feasible reactions, the round-trip accuracy can account for a significant portion of them (see \Cref{fig:ignored_reactions}). The metrics were computed on the USPTO-50k dataset and the "real" feasibility was assessed with USPTO-MIT. Note that the number of ignored feasible reactions is highly underestimated as USPTO-MIT is by no means exhaustive. The space of all reactions is enormous\changed{,} and even simple manipulations of leaving groups of reactants (e.g.\changed{,} changing Cl to Br) are likely to result in a lot of feasible reactions that were not screened in the wet lab before. While all the feasible reactions cannot be included in the dataset directly, the round-trip accuracy can account for some portion of them by leveraging the generalization properties of deep learning. The fact that our round-trip accuracy can account for strictly more feasible reactions than top-k accuracy is essential in the context of drug design, even at the cost of the increased number of non-feasible reactions accounted as feasible. It is because the profit lost caused by the inability to synthesize a drug is drastically higher than the cost of performing an unsuccessful synthesis experiment \cite{burt2017burden}. In \ref{app:drug_design}, we introduce a simplified drug-design pipeline and show that in 2 out of 3 scenarios optimizing round-trip accuracy leads to higher expected profits than optimizing top-k accuracy.


\subsection{Reliability of Round-Trip Accuracy}
\label{sec:reliability}
To assess the reliability of the round-trip accuracy, we want to estimate what percentage of non-feasible reactions the round-trip accuracy will treat as feasible (we call this metric acceptance accuracy). The problem with constructing a set of non-feasible reactions is that they are very rarely reported in the literature. All reaction from USPTO-50k and USPTO-MIT datasets that we consider in that paper consists only of feasible reactions. Therefore, to obtain non-feasible reactions, we assume that for every set of reactants from the USPTO-MIT all of its possible outcomes were reported. Under this assumption, we can create a non-feasible reaction by taking a set of reactants from USPTO-MIT and a product that is not a possible outcome.  We create an initial set of such reactions by applying random forward templates to the $m$ reactants from USPTO-MIT (test split). Then we select a subset $C$ of size $m/10$ of \changed{the} obtained reactions so that all reactions have distinct products and reactants. We call $C$ a set of \textbf{the most challenging} non-feasible reactions. Then, for every reaction $(r, p) \in C$, we gather 9 sets of reactants $r_i$ from USPTO-MIT with possibly high Tanimoto similarity to $r$ and add $(r_i, p)$ reactions to $C$. As a result, for every product $p$ from $C$, we have a set of 10 corresponding reactions that are non-feasible in a non-trivial way. We additionally ensure that the sets of reactants are distinct across the reactions. If a set of reactants were shared between some reactions, then only one of those reactions would be accepted by a forward model, artificially increasing the acceptance accuracy.

The acceptance accuracy of round-trip accuracy (and underlying forward model) is reported in \Cref{fig:rejection_accuracy}. We see that the forward model accurately rejects even the most challenging reactions $C$ obtained by the forward reaction template application. 

\subsection{Round-Trip Accuracy in Drug Design Scenario}
\label{app:drug_design}
In the case of drug discovery, false-negative error rate $\ebeta$ is widely believed to have a much higher influence on the expected investment return than false-positive error rate $\ealpha$ \cite{burt2017burden}. It comes from the fact that $\ebeta$ is roughly proportional to the chances of missing a drug from the experimental pipeline, while $\ealpha$ is roughly proportional to the chances of including a non-drug in the pipeline. And because the costs of an experiment are orders of magnitude lower than the expected profits from developing a drug, it’s very reasonable to decrease the chances of missing a drug (decrease $\ebeta$) at the expense of an increased number of failed experiments (increased $\ealpha$).

In \Cref{app:drug_desing_1}, we analyze the influence of $\ealpha$ and $\ebeta$ of a retrosynthesis model on the expected income of a drug design pipeline. Further, in \Cref{app:error_estimates}, we estimate the $\ealpha$ and $\ebeta$ of models trained to optimize the top-k accuracy (ACC) and top-k round-trip accuracy (RT) metrics. In \Cref{app:results_rt_acc}, we conclude that in 2 out of 3 scenarios\changed{,} optimizing RT leads to better outcomes than optimizing ACC. 

\subsubsection{Drug Design Scenario}
\label{app:drug_desing_1}
In this section, we describe a relationship between $\ealpha$ and $\ebeta$ of the single-step retrosynthesis model and the expected investment return of a simplistic drug design pipeline. 

Let’s assume that we use a single-step retrosynthesis model with error rates $\ealpha$ and $\ebeta$ in the multi-step retrosynthesis planning pipeline (RETRO). For the sake of simplicity, let’s assume that for every molecule there is only one ground-truth synthetic route of length $n=5$. The probability that the ground-truth route can be retrieved by RETRO is roughly equal to $p_1 = (1-\ebeta)^n$ (at every synthetic step\changed{,} we need to be able to recover the ground-truth reaction). If the ground-truth route can be found, RETRO outputs it in $m$ trials with a probability $p_2 = (1-(1 - \ealpha)^n)^m$. The overall probability of retrieving a ground-truth route by RETRO in $m$ trials $p(\ealpha, \ebeta, m) = p_1(1-p_2)$.

Let’s further assume that RETRO proposes synthesis routes for a pool of drug candidates. The synthesis routes are then evaluated in the laboratory and successfully synthesized candidates are pushed further in the drug design pipeline, possibly becoming highly lucrative products. The expected income for $\ealpha$ and $\ebeta$ is calculated as: $max_{m} (P \cdot p(\ealpha, \ebeta, m) - C \cdot m)$. Its value depends mostly on the profit-to-cost ratio $\frac{P}{C}$, which relates the cost of running a single synthesis experiment ($C$) with an expected profit from being able to synthesize a given molecule ($P = P_D \cdot R_D$; where $P_D$ is the profit from developing a drug and the $R_D$ is the ratio of the drugs in the candidates pool). The concrete values of $P$ and $C$ are hard to estimate globally. For this reason, we evaluate three scenarios with  $\frac{P}{C}$ $\in \{10, 100, 1000 \}$. The middle value was calculated based on the estimation of drug profit $P_D=2e9$ from \cite{burt2017burden}, drug ratio $R_D=1e-5$ from \cite{sun202290} (there are $1e4$ compounds in the screening phase and only one of them will be a drug with a 10\% chance), and synthesis cost $C=2e2$ from \cite{koziarski2024rgfn}.

\begin{figure*}[t]
\begin{center}
\centerline{\includegraphics[width=\linewidth]{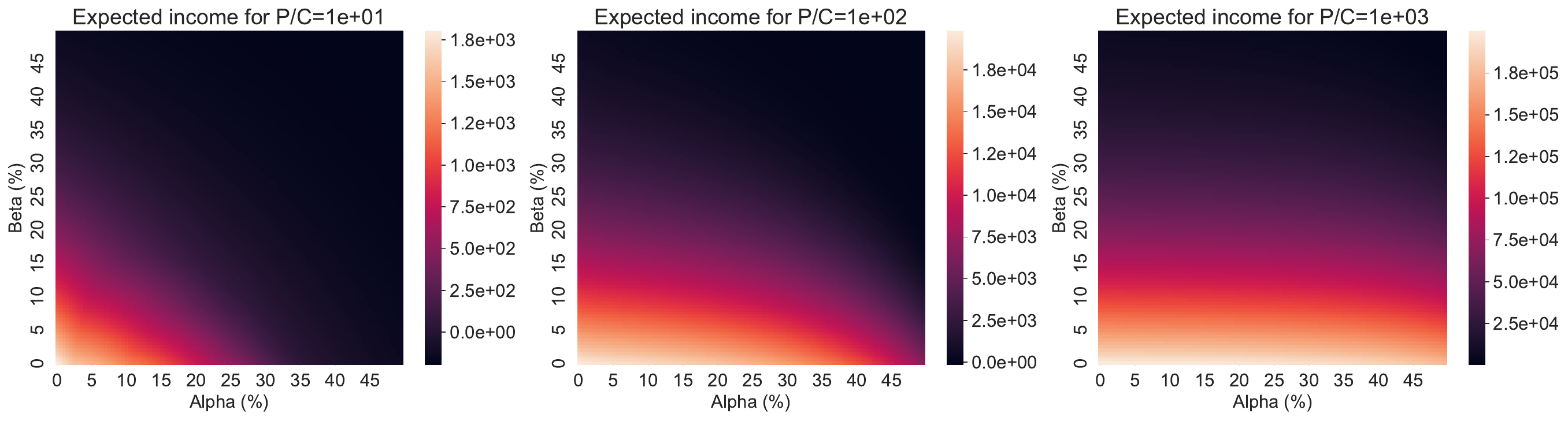}}
\caption{The estimated income of drug design pipeline as a function of false positive ($\ebeta$) and false negative ratio ($\ealpha$) of a single-step retrosynthesis model. For scenarios with moderate and high \changed{profit-to-cost} ratio (P/C), the $\ebeta$ is significantly more important than $\ealpha$. }
\label{fig:drug_design}
\end{center}
\vskip -0.2in
\end{figure*}

We plot the expected income as \changed{a} function of $\ealpha$ and $\ebeta$ in \Cref{fig:drug_design} for those three scenarios. \Cref{fig:drug_design} recalls the common belief that $\ebeta$ influences the expected income much more than $\ealpha$.

\subsubsection{Estimates of Error Rates of ACC and RT}
\label{app:error_estimates}
To assess $\beta$ of ACC, we gathered all reactions from USPTO-MIT that: 1) share a product with some ground-truth reaction from USPTO-50k, and 2) are not in USPTO-50k. There are 2532 such reactions, suggesting $\ebeta \approx 5\%$ for ACC. On the other hand, the forward model underlying RT was able to account for 1294/2532 of them\changed{,} suggesting $\ebeta \approx 2.5\%$. 
We assume that $\ealpha=0$ for ACC. \Cref{fig:rejection_accuracy} suggests that $\ealpha \approx 10\%$ for RT as the underlying forward model rejects ~90\% of the most tricky negative reactions. This number also reflects the top-1 forward prediction accuracy declared in the Chemformer paper \cite{irwin2022chemformer}.

\subsubsection{Results for RT- and ACC-optimal models}
\label{app:results_rt_acc}
We report the expected income for the ACC-optimal model (an idealized model that obtains perfect ACC; by definition it has $\ealpha=0\%$, $\ebeta=5\%$) and RT-optimal model ($\ealpha=10\%$, $ \ebeta=2.5\%$) for scenarios with different profit-to-ratio (P/C) magnitudes in \Cref{tab:rt_acc}. We observe that in 2 out of 3 scenarios\changed{,} optimizing RT leads to better outcomes than optimizing ACC. 

\begin{table*}[!ht]
    \centering
    \caption{The expected return of investment in drug design pipeline for different profit-to-cost ratio P/C (columns) and single-step synthesis models optimized with respect to top-k accuracy (ACC) and round-trip accuracy (RT). We observe that in 2 out of 3 scenarios\changed{,} optimizing RT leads to better outcomes than optimizing ACC. }
    \label{tab:rt_acc}
    \begin{tabular}{@{}lccc@{}}
    \toprule
        model & 10 & 100 & 1000 \\ 
        \midrule
        RT-optimal model & 1.1e3 & \textbf{1.6e4} & \textbf{1.7e5} \\ 
        ACC-optimal model & \textbf{1.3e3} & 1.5e4 & 1.5e5 \\ 
        \bottomrule
    \end{tabular}
\end{table*}

\subsection{Generalization of Round-Trip Accuracy}
\label{sec:extension}
Standard round-trip accuracy formulation counts a reaction as feasible if the forward reaction prediction model outputs the product from the reaction given its reactants as input. Then, the feasibility is a binary feature of the reaction. In \ref{app:rfm}, we propose a generalized version of round-trip accuracy that can deal with the probabilistic notion of feasibility\changed{,} allowing the usage of a wider class of feasibility prediction models. We show that our RetroGFN can greatly improve this metric.

\section{Conclusions}
\label{sec:conclusions}
In this paper, we provided empirical arguments for the importance of reporting the round-trip accuracy in the single-step retrosynthesis model evaluation. Leveraging the GFlowNet framework\changed{,} which is designed for tasks where plenty of sensible solutions are desired, we developed a RetroGFN model that achieves competitive results on top-k accuracy and on the top-k round-trip accuracy. We discuss the limitations of the paper in the \ref{app:limitations}.

\section{Future Work}
\label{sec:future_work}
There are two main directions of future work on RetroGFN. The first is to mitigate its current limitations including an improvement of the inference, adaptation of more robust feasibility models, or tuning the RetroGFN's design choices. The second direction is to develop novel single-step retrosynthesis models that are guided by a feasibility model. An especially interesting idea is to design a template-free version of RetroGFN that would unlock the full potential of the GFlowNets framework to generate diverse reactions.

\section*{Acknowledgements} The research of P. Gaiński was supported by the National Science Centre (Poland), grant no. 2022/45/B/ST6/01117. The research of M. Śmieja was supported by the National Science Centre (Poland), grant no. \changed{2023/49/B/ST6\\/01137}. The research of J. Tabor was supported by the National Science Centre (Poland), grant no. 2021/41/B/ST6/01370. The research of M. Koziarski was supported by funding from CQDM Fonds d'Accélération des Collaborations en Santé (FACS) / Acuité Québec and Genentech. We gratefully acknowledge Poland's high-performance Infrastructure PLGrid (ACK Cyfronet Athena, HPC) for providing computer facilities and support within computational grant no PLG/2023/016550. For the purpose of Open Access, the author has applied a CC-BY public copyright license to any Author Accepted Manuscript (AAM) version arising from this submission.

\section*{Credit Statement} \textbf{Piotr Gaiński}: Conceptualization, Methodology, Investigation, Software, Writing - Original Draft. \textbf{Michał Koziarski}: Conceptualization, Methodology, Writing - Review \& Editing. \textbf{Krzysztof Maziarz}: Conceptualization, Methodology, Writing - Review \& Editing. \textbf{Marwin Segler}: Conceptualization, Methodology, Writing - Review \& Editing. \textbf{Jacek Tabor}: Conceptualization, Writing - Review \& Editing. \textbf{Marek Śmieja}: Supervision, Conceptualization, Writing - Review \& Editing.



\bibliographystyle{elsarticle-num}
\bibliography{bib}

\newpage
\appendix
\onecolumn

\section{RetroGFN Details}
\label{app:retro_gfn_details}
All neural networks in RetroGFN used the same hidden dimension $h=200$. To obtain initial node and edge features for products, we used featurization from \cite{kearnes2016molecular}. For the reactant pattern, we used the same edge featurization and a custom node featurization that accounted for atom type, degree, aromaticity, whether the atom was mapped in the original template, relative charge difference between product and reactant atom in the original template, and analogous implicit hydrogen difference. The node features for both products and reactant patterns were enriched with random walk positional encoding 
of size $\mathrm{n\_random\_walk}=16$.

Product node encoder $\gnn_1$ consists of $\mathrm{num\_layer\_1}=4$ layers of the MPNN convolution \cite{gilmer2017neural} and one Transformer layer with $\mathrm{num\_heads}=8$. The reactant pattern encoder differs only in the number of layers $\mathrm{num\_layer\_2}=3$. Multi-layer perceptrons $\MLP_1, \MLP_2, \MLP_3$ had one hidden layer (with hidden dimension $h$) and used the GeLU activation function. 

During training, we used a combination of three sampling methods: 1) standard exploratory sampling from the forward policy $P_F$ with some $\epsilon$ probability of taking random actions, 2) backward sampling from replay buffer \cite{shen2023towards,fedus2020revisiting,jain2022biological}, and 3) backward sampling from the dataset $D$. Backward sampling starts with a terminal state and samples the trajectory in the backward direction using the backward policy. 

During the training probability of taking random action in the forward policy was set to $\epsilon=0.05$, the number of sampled forward trajectories in the batch was $\mathrm{n\_forward}=16$, and the analogous numbers for backward dataset trajectories and backward replay buffer trajectories were $\mathrm{n\_dataset}=96$ and $\mathrm{n\_replay}=16$. The model was trained with Adam optimizer \cite{kingma2014adam} with a learning rate $lr=0.0005$ (with other parameters set to default values in the torch implementation) for $\mathrm{n_iterations}=25000$ iterations. In the evaluation, the forward policy temperature was set to $\alpha=0.7$. During the inference, we sampled $K \cdot N$ trajectories to accurately estimate the reaction probability. For USPTO-50k, we set $K=20$ while, due to limited computational resources, we set $K=10$ for USPTO-MIT. 

All the hyperparameters were chosen manually based on the top-k accuracy and round-trip accuracy estimated on the USPTO-50k validation split.

\section{Generalization of Round-Trip Accuracy}
\label{app:rfm}
We propose a generalization of round-trip accuracy that allows to use of a wider class of machine-learning models to assess the reaction feasibility. We call this metric Feasible Thresholded Count (FTC).
For a single product, the top-k FTC value denotes the percentage of feasible reactions among \changed{the} top $k$ reactions returned by the model. The feasibility is estimated with an auxiliary model $\RFM$ described \changed{further} in this section. The exact formula for top-k \ftc{} calculated on a product $p$ and retrosynthesis model $g$ is given by:


\begin{equation}
    \mathrm{F_{\ftc}}(g, p, k) = \frac{1}{k} \sum_{i=1}^k \mathbb{1}[\RFM(g(p)_i) \geq t],
\end{equation}
where $\RFM(g(p)_i) \in [0, 1]$ is the output of the reaction feasibility model for the $i$-th reaction proposed by $g$, and $t$ is a feasibility threshold given by the user. We assume that $\RFM(x) = 1$ for reaction $x \in D$. We report the top-k \ftc{} for the entire dataset $D$, which can be written as $\ftc(g, k) = \frac{1}{n}\sum_{i}^n\mathrm{F_{\ftc}}(g, p_i, k)$. 

\subsection{Reaction Feasibility Model (RFM)}
\label{sec:feasibility_model}
The Reaction Feasibility Model (RFM) is a model that takes reaction $x$ as an input and outputs its feasibility - probability that the reaction is feasible: $\RFM(x) \in [0, 1]$. In this paper, we develop an RFM baseline that can be used as a benchmark in future work.

\subsubsection{Architecture} Our RFM implementation consists of two Graph Neural Networks (GNN) layers with a Transformer \cite{vaswani2017attention} layer and attention pooling at the top that creates product and reactant embeddings\changed{,} which are then concatenated and fed into the $\MLP$ layer.

\subsubsection{Checkpoints for USPTO-50k} To train the model, we augmented the USPTO-50k dataset with negative (non-feasible) reactions using two methods: 1) application of existing forward templates to obtain a novel product from existing reactants, 2) swapping a product in the reaction with another product that is similar to the original one in terms of Tanimoto similarity. Such an approach ensured that the generated negative reactions are not trivially unfeasible (they use an existing template and/or the product is not strikingly different from the reactants), but still are very unlikely to occur in reality (the original reactants were reported to return a different product). We obtained a reaction feasibility dataset with a negative-to-positive ratio of 5:1.  We trained two distinct checkpoints of feasibility models: RFM-Train-50k and RFM-Eval-50k. The RFM-Train was trained only on the train split of the reaction feasibility dataset and was then used to calculate the reward in the RetroGFN during the training. 

\subsection{Experiments}
We trained the RetroGFN using the RFM-Train model as a feasibility proxy and compared it on top-k accuracy and our \ftc{} metric. We used the same hyperparameters as in \ref{app:retro_gfn_details}, but with $\mathrm{n\_dataset}=80, \mathrm{n\_replay}=16, \mathrm{n\_forward}=32$ and $\beta=12$. The results (Tables \ref{tab:ftc_50k_2}, \ref{tab:standard_50k_2}, \ref{tab:ftc_mit_2} and \ref{tab:standard_mit_2}) mimic the ones from the main paper: our RetroGFN outperforms the model on \ftc{} metric while obtaining competitive results on the standard top-k accuracy. The experiments show that our RetroGFN can leverage any machine-learning feasibility proxy. We believe that training a reliable and powerful feasibility proxy is a promising direction for future work.

\begin{table*}[!ht]
    \centering
    \caption{Top-k \ftc{} results on USPTO-50k along with the standard deviation for threshold $t=0.9$. The best results in every column are \textbf{bolded}. We observe that for $k>1$ our RetroGFN consistently outperforms other methods.}
    \label{tab:ftc_50k_2}
    \begin{tabular}{@{}lcccccc@{}}
    \toprule
        method & top-1 & top-3 & top-5 & top-10 & top-20 & top-50 \\ \midrule
        GLN & 74.0 & 57.4 & 49.0 & 38.1 & 28.4 & 16.4 \\ 
        MEGAN & 70.4 & 54.2 & 46.7 & 37.3 & 28.6 & 18.5\\ 
        MHNreact & 72.0 & 54.1 & 45.5 & 35.0 & 25.6 & 14.3 \\ 
        LocalRetro & 73.4 & 57.2 & 49.8 & 40.0 & 31.4 & 20.1 \\ 
        RetroKNN & 72.0 & 54.7 & 47.0 & 35.7 & 22.5 & 9.5 \\ 
        RootAligned & \textbf{75.9} & 57.2 & 49.7 & 40.3 & 31.4 & 16.1 \\ 
        Chemformer & 74.9 & 41.5 & 28.5 & 15.7  & 8.2 & 3.4 \\ 
        RetroGFN & 72.4 & \textbf{61.0} & \textbf{54.1} & \textbf{46.0} & \textbf{38.9} & \textbf{29.1} \\ \bottomrule
    \end{tabular}
\end{table*}

\begin{table}[!ht]
    \centering
    \caption{
        Top-k accuracy results on USPTO-50k. The numbers in columns denote $k$ values. The best results in every column are \textbf{bolded}. We observe that for $k>3$ our RetroGFN achieves competitive results.
    }
    \label{tab:standard_50k_2}
    \begin{tabular}{lccccccc}
    \toprule
        method & MRR & top-1 & top-3 & top-5 & top-10 & top-20 & top-50 \\ \midrule
        GLN & 0.6509 & 52.4 & 74.6 & 81.2 & 88.0 & 91.8 & 93.1 \\ 
        MEGAN & 0.6226 & 48.7 & 72.3 & 79.5 & 86.7 & 91.0 & 93.5 \\ 
        MHNreact & 0.6356 & 50.6 & 73.1 & 80.1 & 86.4 & 90.3 & 92.6 \\ 
        LocalRetro & 0.6565 & 51.5 & 76.5 & 84.3 & 91.0 & \textbf{94.9} & \textbf{96.7} \\ 
        RetroKNN & 0.6834 & 55.3 & 77.9 & 85.0 & \textbf{91.5} & 91.6 & 96.6 \\ 
        RootAligned & \textbf{0.6886} & \textbf{56.0} & 79.1 & \textbf{86.1} & 91.0 & 93.3 & 94.2 \\ 
        Chemformer & 0.6312 & 55.0 & 70.9 & 73.7 & 75.4 & 75.9 & 76.0 \\ 
        
        RetroGFN & 0.6144 & 46.9 & 72.2 & 80.0 & 87.8 & 91.9 & 94.7 \\ \bottomrule
    \end{tabular}
\end{table}

\begin{table*}[!ht]
    \centering
    \caption{Top-k \ftc{} results on USPTO-MIT along with the standard deviation for threshold $t=0.9$. The best results in every column are \textbf{bolded}. We observe that for $k>1$ our RetroGFN consistently outperforms other methods.}
    \label{tab:ftc_mit_2}
    \begin{tabular}{@{}lcccccc@{}}
    \toprule
        method & top-1 & top-3 & top-5 & top-10 & top-20 & top-50 \\ \midrule
        GLN & 74.8 & 64.2 & 58.3 & 49.7 & 40.6 & 27.6 \\ 
        MEGAN & 72.9 & 61.5 & 55.6 & 47.2 & 38.8 & 27.1 \\ 
        MHNreact & 73.3 & 61.7 & 55.2 & 46.1 & 36.6 & 22.7 \\ 
        LocalRetro & 76.3 & 65.4 & 59.4 & 51.3 & 42.9 & 30.2 \\ 
        RootAligned & \textbf{77.6} & 66.0 & 60.0 & 51.7 & 42.4 & 22.8 \\ 
        Chemformer & 73.8 & 47.1 & 34.2 & 20.2 & 11.0 & 4.6 \\ 
        RetroKNN & 75.0 & 62.9 & 56.1 & 45.1 & 29.8 & 12.8 \\ 
        RetroGFN & 77.4 & \textbf{69.1} & \textbf{64.5} & \textbf{58.1} & \textbf{51.7} & \textbf{40.7} \\ \bottomrule
    \end{tabular}
\end{table*}

\begin{table*}[!ht]
    \centering
    \caption{
        Top-k accuracy results on USPTO-MIT. The numbers in columns denote $k$ values. The best results in every column are \textbf{bolded}. We observe that for $k>3$ our RetroGFN achieves competitive results.
    }
    \label{tab:standard_mit_2}
    \begin{tabular}{lccccccc}
    \toprule
        method & MRR & top-1 & top-3 & top-5 & top-10 & top-20 & top-50 \\ \midrule
        GLN & 0.4480 & 35.6 & 51.5 & 56.5 & 61.6 & 64.2 & 65.3 \\ 
        MEGAN & 0.4498 & 35.3 & 52.0 & 57.6 & 62.6 & 65.8 & 68.1 \\ 
        MHNreact & 0.4451 & 35.3 & 51.3 & 56.4 & 60.9 & 63.7 & 65.2 \\ 
        LocalRetro & 0.4636 & 36.0 & 54.2 & 59.9 & \textbf{65.1} & \textbf{67.9} & \textbf{69.7} \\ 
        RetroKNN & 0.4491 & 34.9 & 52.5 & 58.2 & 63.5 & 65.3 & 65.5 \\
        RootAligned & \textbf{0.4838} & \textbf{38.9} & \textbf{55.6} & \textbf{60.6} & 65.2 & 67.7 & 68.8 \\
        Chemformer & 0.4362 & 37.8 & 49.1 & 51.2 & 52.5 & 52.9 & 52.9 \\ 
        RetroGFN & 0.4375 & 33.1 & 51.3 & 57.5 & 63.3 & 66.7 & 68.9 \\ \bottomrule
    \end{tabular}
\end{table*}

\section{Ablations}
\label{sec:ablations}
In this section, we study a simple model-agnostic way of leveraging the Chemformer-Train to maximize the results of the round-trip accuracy metric. The idea is to filter the results that are not backtranslated by the Chemformer-Train model during the evaluation. Tables \ref{tab:standard_ablation} and \ref{tab:round_trip_ablation} show that such an approach significantly improves the round-trip accuracy results, but with the \changed{cost} of a drastic decrease \changed{in} a standard top-k accuracy, especially for larger values of $k$. 

\begin{table}[!ht]
    \centering
    \caption{
        Top-k accuracy results on USPTO-50k for models that use the Chemformer-Train filtering. The best results in every column are \textbf{bolded}. We observe that the performance of all the models is significantly degraded, especially for larger values of $k$.
    }
    \label{tab:standard_ablation}
    \begin{tabular}{@{}lcccccc@{}}
    \toprule
    method              & top-1 & top-3 & top-5 & top-10 & top-20 & top-50 \\ \midrule
    MEGAN + filter      & 48.4  & 70.1  & 75.9  & 81.2   & 83.3   & 83.8   \\
    LocalRetro + filter & \textbf{49.6}  & \textbf{71.9}  & \textbf{78.3}  & \textbf{83.2}   & \textbf{85.5}   & \textbf{86.3}   \\
    RetroGFN + filter   & 46.9  & 68.8  & 75.3  & 81.3   & 83.9   & 85.4   \\ \midrule
    MEGAN               & 48.7  & 72.3  & 79.5  & 86.7   & 91.0   & 93.5   \\
    LocalRetro          & \textbf{51.5}  & \textbf{76.5}  & \textbf{84.3}  & \textbf{91.0}   & \textbf{94.9}   & \textbf{96.7}   \\
    RetroGFN            & 49.2  & 73.3  & 81.1  & 88.0   & 92.2   & 95.3   \\ \bottomrule
    \end{tabular}
\end{table}

\begin{table}[!ht]
    \centering
    \caption{
        Top-k round-trip accuracy results on USPTO-50k for models that use the Chemformer-Train filtering. The best results in every column and group are \textbf{bolded}. We observe that the performance of all the models is significantly improved, \changed{however, at the price} of degrading the top-k accuracy results.
    }
    \label{tab:round_trip_ablation}
    \begin{tabular}{@{}lcccccc@{}}
    \toprule
    method              & top-1 & top-3 & top-5 & top-10 & top-20 & top-50 \\ \midrule
    MEGAN + filter      & 95.0  & 92.8  & 90.8  & 85.3   & 71.5   & 37.0   \\
    LocalRetro + filter & \textbf{96.6}  & 94.8  & 93.5  & 90.3   & 82.7   & 48.8   \\ 
    RetroGFN + filter   & 96.1  & \textbf{94.9}  & \textbf{93.9}  & \textbf{91.4}   & \textbf{86.4}   & \textbf{57.2}   \\ \midrule
    MEGAN               & 87.0  & 80.7  & 76.5  & 68.5   & 58.2   & 41.7   \\
    LocalRetro          & \textbf{93.0}  & 87.6  & 84.6  & 79.3   & 71.0   & 54.2   \\
    RetroGFN            & 91.7  & \textbf{88.2}  & \textbf{85.6}  & \textbf{81.1}   & \textbf{74.8}   & \textbf{63.5}   \\ \bottomrule
    \end{tabular}
\end{table}




\section{Limitations and Discussion}
\label{app:limitations}
This section briefly discusses the limitations of the paper.
\subsection{Round-trip Accuracy}
The main limitation of the top-k round-trip accuracy is that it relies on the forward reaction prediction model\changed{,} which suffers \changed{from} both false negative and false positive errors. However, we believe that there is an inherent epistemic uncertainty within the notion of feasibility (we cannot screen all the reactions)\changed{,} and any sensible retrosynthesis metric will have some portion of false negatives (it will not take all feasible reactions into account). In comparison to top-k accuracy, our round-trip accuracy has a strictly lower number of false negatives, while keeping false positives \changed{at} a decent level. We believe that the round-trip will benefit from the further improvements of the forward reaction prediction model and we leave it for future work.

\subsection{RetroGFN}
\subsubsection{Top-k Accuracy} The main limitation of our RetroGFN method is its results on top-k accuracy for $k < 5$. At first glance, it looks like a trade-off necessary to achieve excellent results on the round-trip accuracy. We argue that it may be caused by two things: 1) other hyperparameters of the model are not optimal for top-k accuracy, 2) the GFlowNet framework struggles with \changed{a} spiky reward function, and 3) the parametrization of the composition process is sub-optimal. It is possible that further refinements of the method could improve the results.

\subsubsection{Leveraging Chemformer-Train} The fact that RetroGFN leverages the Chemformer-Train checkpoint can be seen as an unfair advantage because a similar Chemformer-Eval model is used in the round-trip accuracy computation. However, we think that fairness comes from the fact that all models use the same data splits for training or evaluation. The models differ in the way they learn from the training data\changed{,} and leveraging the Chemformer-Train is yet another way of learning. It does not inject any new knowledge that cannot be extracted from the training data. Once the round-trip accuracy metric is established, it becomes reasonable to optimize it using Chemformer-Train. Moreover, we believe that Chemformer-Eval and Chemformer-Train are expected to be similar because they have similar goals: 1) to extract as much information from the train and test split as possible, and 2) to extract as much information from the train split as possible. It is sensible then that they share architecture. The difference should come from the data split used for training.

\section{Computational Resources}
\label{app:resources}
We ran all the experiments on Nvidia V100 and A100 GPUs. The training of our model takes no more than 48h per checkpoint. When experimenting with the architecture and different feasibility proxy models, we trained no more than 100 checkpoints. For all the baselines, we used already trained checkpoints and only evaluated them on USPTO-50k and USPTO-MIT. The evaluation time depends on the model, but in total, it took no more than 400 GPU hours. It gives the upper bound of 5200 GPU hours for the total \changed{experiments'} costs.

\end{document}